\documentclass[a4paper]{report}
\usepackage[utf8]{inputenc}
\usepackage[T1]{fontenc}
\usepackage{RJournal}
\usepackage{amsmath,amssymb,array}
\usepackage{booktabs}

\providecommand{\tightlist}{%
  \setlength{\itemsep}{0pt}\setlength{\parskip}{0pt}}

\begin{document}

\sectionhead{Contributed research article}
\volume{XX}
\volnumber{YY}
\year{20ZZ}
\month{AAAA}

\begin{article}
\title{chemmodlab: A Cheminformatics Modeling Laboratory for Fitting and
Assessing Machine Learning Models}
\author{by Jeremy R. Ash, Jacqueline M. Hughes-Oliver}

\maketitle

\abstract{%
The goal of \pkg{chemmodlab} is to streamline the fitting and assessment
pipeline for many machine learning models in R, making it easy for
researchers to compare the utility of new models. While focused on
implementing methods for model fitting and assessment that have been
accepted by experts in the cheminformatics field, all of the methods in
\pkg{chemmodlab} have broad utility for the machine learning community.
\pkg{chemmodlab} contains several assessment utilities including a
plotting function that constructs accumulation curves and a function
that computes many performance measures. The most novel feature of
\pkg{chemmodlab} is the ease with which statistically significant
performance differences for many machine learning models is presented by
means of the multiple comparisons similarity plot. Differences are
assessed using repeated k-fold cross validation where blocking increases
precision and multiplicity adjustments are applied.
}

\section{Introduction}\label{introduction}

It is now commonplace for researchers across a variety of fields to fit
machine learning models on complex data to make predictions. The
complexity of these data (e.g., large number of features, non-linear
relationships with the response) often means it is difficult to
determine a priori what machine learning modeling routine and what
descriptors (also known as features, predictors, or covariates) will
result in the best performance. A common approach to this problem is to
fit many descriptor set and modeling routine (D-M) combinations, and
then compute measures of prediction performance for held out data to
choose a D-M combination by assessing relative performance.

Often in a particular domain, there are only a few modeling routines
that are widely accepted, and researchers tend to use these methods
exclusively. Unfortunately, this will not always work well for every
data set and researchers might learn from other fields where different
modeling methods tend to be more succesful. There are a myraid of
modeling methods implemented in R that may be worthwhile for researchers
to try (see \citet{Hastie2009} and \citet{Kuhn2013} for an overview of
these methods). However, the lack of knowledge of the syntactic minutiae
and statistical methodology that is required to fit and compare
different modeling routines in R often prohibits users from attempting
them.

\CRANpkg{chemmodlab} currently implements 13 different machine learning
models. Fitting the entire suite of models requires little user
intervention -- all models are fit with a single command. Sensible
defaults for tuning parameters are set automatically, but users may also
tune models outside of \pkg{chemmodlab} (in \CRANpkg{caret}
\citep{McCollum2009, Kuhn2016}, for example) and then manually set the
tuning parameters.

While the package \textbf{caret} has a similar goal, the fitting of many
D-M combinations and the determination of statistically significant
performance differences still requires some knowledge of R programming
and statistical methods. \pkg{chemmodlab} further automates this process
and provides several plotting utilities for easily assessing the
results, lowering the barrier of entry for researchers unfamiliar with
these methods.

One motivation for this package was the observation that once
performance measures are computed for several different D-M
combinations, researchers often do not consider the randomness and
uncertainity involved in obtaining the observed performance measures. If
one model has prediction performance that is marginally better than
another, it is tempting to claim improvement. However, \emph{very
slight} changes in the original data set or in how assessment was
conducted could have led to a reversal of observed performance. By
accounting for the inherent uncertainity in data collection and model
assessment, a stronger and more defensible claim can be made about
differences in prediction performance. For example, a carefully
constructed confidence interval that does not contain zero for the
difference in performance measures between two D-M combinations would
reliably identify significant differences between the two D-M
combinations, even after accounting for uncertainty.

Figure \ref{fig:CombineSplits_auc} shows an example of this. Many
classification models have been fit to two different descriptor sets to
predict a binary response variable. There are a total of 18 D-M
combinations to be compared. The D-M combinations were assessed using
repeated 10-fold cross validation and the area under the receiver
operating characteristic curve (AUC) performance measure. A multiple
comparisons (MCS) plot visualizes the differences in model performance
(Figure \ref{fig:CombineSplits_auc}). The descriptor sets will be
discussed in detail later on, but for now, it is sufficient to say that
the Pharmacophore descriptors are far more interpretable than the Burden
Number descriptors. The Burden Numbers-Random Forest (RF) combination is
the best performing D-M combination (AUC: .76). However, the
Pharmacophore-Least Angle Regression (LAR) combination (AUC: .71)
involves a highly interpretable linear model with a subset of the
Pharmacophore descriptors selected. This .05 difference is small and
without additional investigations it is unclear whether it is
statistically significant.

By performing multiple cross validation splits and using these splits as
a blocking factor to improve precision, \pkg{chemmodlab} is able to test
for statistical significance of performance measure differences and
visualizes these results in a manner that can be easily interpreted by
the user. The question this addresses is: if the experiment were
repeated with changes to the training and/or test set, would the best
performing model still be the best? Again referring to Figure
\ref{fig:CombineSplits_auc}, the MCS plot indicates that a significance
level of 0.01 leads to the conclusion that the two aformentioned D-M
combinations (Burden Numbers-RF, and Pharmacophores-LAR) are equivalent,
and hence the more interpretable model may be selected for further
investigations. This inference has been multiplicity-adjusted for the
\({18 \choose 2} = 153\) pairwise comparisons that were made.

\begin{Schunk}
\begin{figure}

\includegraphics{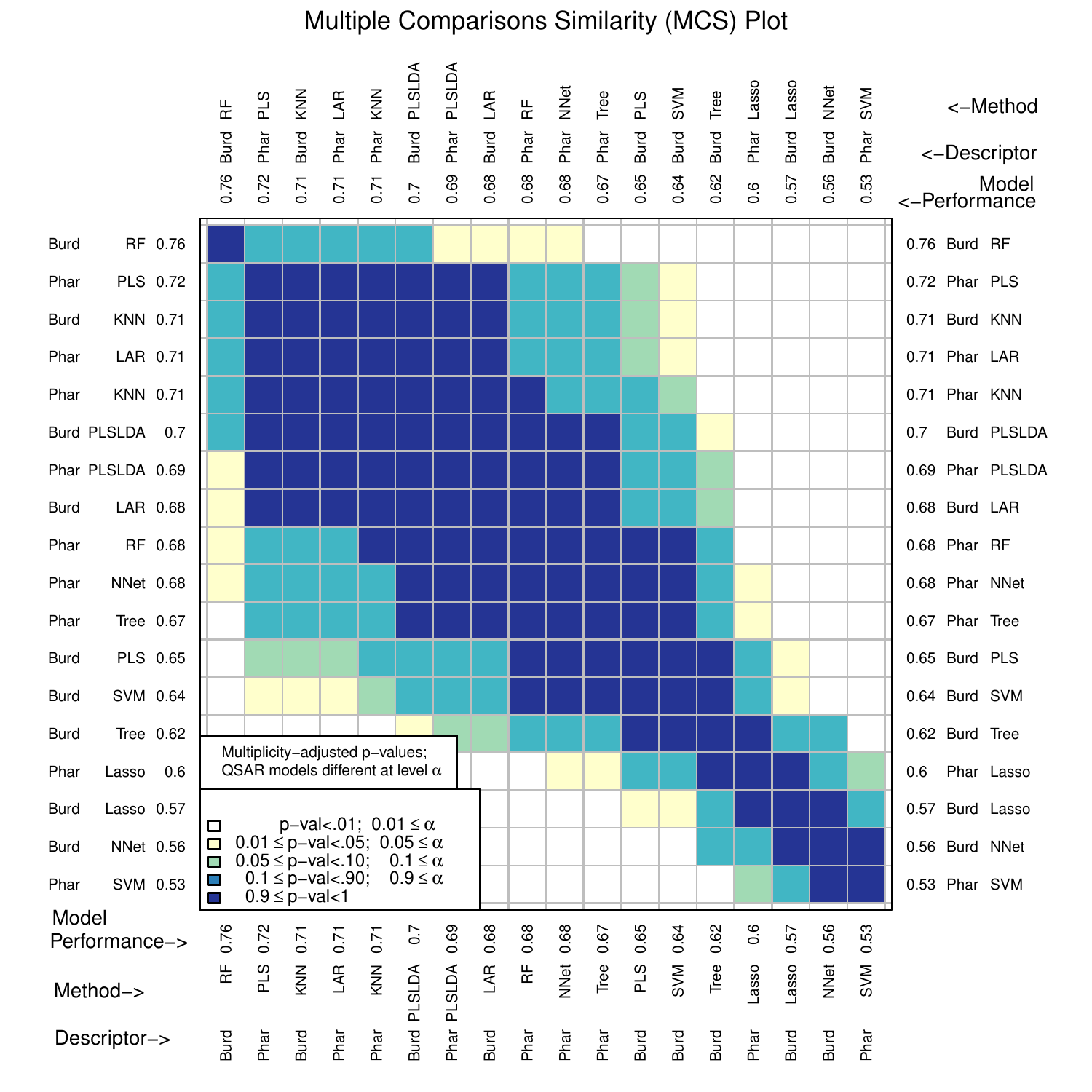} \hfill{}

\caption{\label{fig:CombineSplits_auc}MCS plot using area under the receiver operating characteristic curve as the performance measure.}\label{fig:CombineSplits_auc}
\end{figure}
\end{Schunk}

\pkg{chemmodlab} is a re-creation and extension of the former ChemModLab
webserver \citep{Hughes-Oliver2011}. Some notable extensions to the
previous software are:

\begin{itemize}
\tightlist
\item
  \pkg{chemmodlab} has been redesigned so that it is usable in the R
  environment. Models are fit with a simple command, producing an object
  that can be easily passed to plotting and performance assessment
  functions;
\item
  much more control over the model fitting and assessment functions.
  There are now many arguments for customizing the procedures and the
  output they provide, including the appearance of the plots generated.
  As an example, \pkg{chemmodlab} now has support for user supplied
  tuning parameters so that models can be better fit to the data at
  hand;
\item
  and many new model performance measures for classification and
  regression have been implemented.
\end{itemize}

\pkg{chemmodlab} is organized into two successive components: (1) model
fitting, which is primarily conducted via function \code{ModelTrain} and
(2) model assessment, which is conducted via function
\code{CombineSplits}.

\section{An illustrative example}\label{an-illustrative-example}

\subsection{Preparing the data}\label{preparing-the-data}

We will use a cheminformatics data set to illustrate a typical analysis
pipeline in \pkg{chemmodlab}. In cheminformatics, chemists often build
machine learning models that relate a chemical compound's structure to
an observed endpoint. These models are collectively referred to as
quantitative structure-activity relationship (QSAR) models. See
\citet{Cherkasov2014} for an excellent review of the ways in which these
models have played a crucial role in the drug discovery process. Often
the ``endpoint'' (or response variable) is a measure of a compound's
biological activity, which may be either binary, indicating
active/inactive, or a continuous measure of binding affinity for a
target protein. Chemical descriptors represent various levels of
organization of a chemical's structure. See \citet{Cherkasov2014} for an
overview of commonly used chemical descriptors.

The data we will analyze is from a cytotoxicity assay conducted by the
Scripps Research Institute Molecular Screening Center. There are 3,311
compounds, 50 of which are active. Visit
\url{http://pubchem.ncbi.nlm.nih.gov/assay/assay.cgi?aid=364} for more
details, though this data set has been slightly edited since we
downloaded the version used here.

For completeness, the preprocessing of the data set analyzed will be
shown. First, the response variable and molecule IDs are read in from
file.

\begin{Schunk}
\begin{example}
aid364 <- read.csv("AID_364_response.csv")
head(aid364)
\end{example}
\begin{Soutput}
#>        CID Outcome
#> 1  5388992       1
#> 2  5388983       1
#> 3   663143       1
#> 4    10607       1
#> 5  5388972       1
#> 6 11970251       1
\end{Soutput}
\end{Schunk}

Next, two descriptor sets are added to the data frame. Both of these
sets were computed using the software PowerMV - see \citet{Liu} for more
information. The first set of 24 continuous descriptors are a
modification of the Burden number descriptors \citep{Burden1989}.

\begin{Schunk}
\begin{example}
desc.lengths <- c()
d <- read.csv("BurdenNumbers.csv")
head(d[, 1:5])
\end{example}
\begin{Soutput}
#>        Row WBN_GC_L_0.25 WBN_GC_H_0.25 WBN_GC_L_0.50 WBN_GC_H_0.50
#> 1  5388992      -2.40010       1.98339      -2.52864       2.50835
#> 2  5388983      -2.40010       1.98240      -2.52868       2.50398
#> 3   663143      -2.41650       1.32890      -2.53910       2.05778
#> 4    10607      -2.38337       2.17677      -2.52643       2.33232
#> 5  5388972      -2.29039       1.97468      -2.41743       2.46177
#> 6 11970251      -2.29039       2.22488      -2.41748       2.56161
\end{Soutput}
\begin{example}
aid364 <- merge(aid364, d, by.x = "CID", by.y = "Row")
desc.lengths <- c(desc.lengths, ncol(d[-1]))
\end{example}
\end{Schunk}

The number of descriptors in each descriptor set are also stored, as
this information will be used to parse the data frame later on. The
second set contains 147 binary descriptors, indicating the
presence/absence of ``pharmacophore'' features, described in more detail
in \citet{Liu}.

\begin{Schunk}
\begin{example}
d <- read.csv("Pharmacophores.csv")
head(d[, 1:6])
\end{example}
\begin{Soutput}
#>        Row NEG_01_NEG NEG_02_NEG NEG_03_NEG NEG_04_NEG NEG_05_NEG
#> 1  5388992          0          0          0          0          0
#> 2  5388983          0          0          0          0          0
#> 3   663143          0          0          0          0          0
#> 4    10607          0          0          0          0          0
#> 5  5388972          0          0          0          0          0
#> 6 11970251          0          0          0          0          0
\end{Soutput}
\begin{example}
aid364 <- merge(aid364, d, by.x = "CID", by.y = "Row")
desc.lengths <- c(desc.lengths, ncol(d[-1]))
\end{example}
\end{Schunk}

A subset of this data set, containing all 50 active compounds and an
additional 450 compounds, is included in \pkg{chemmodlab}.

\section{Model fitting}\label{model-fitting}

\subsection{\texorpdfstring{The \code{ModelTrain}
function}{The  function}}\label{the-function}

For the model fitting component of \pkg{chemmodlab}, the primary
function is \code{ModelTrain}, which fits a series of classification or
regression models to a data set.

Function \code{ModelTrain} takes as input a data frame with an
(optional) ID column, a response column, and descriptor columns. We have
processed the \code{aid364} data set so that it follows this format. The
specification of an ID column allows users to easily match response
predictions to their observation IDs in the \pkg{chemmodlab} output.

\pkg{chemmodlab} can currently handle responses that are continuous or
binary (represented as a numeric vector with 0 or 1 values). Assessment
assumes preference for large response values. Users can specify which
columns in the data frame they would like to consider as distinct
descriptor sets. At the moment, the response and descriptors may only be
binary or continuous, though we are currently working on support for
categorical variables of more than two levels.

For our example, we previously stored the number of descriptors in each
descriptor set in an integer vector named \code{desc.lengths}, with the
ordering of the integers matching the order of the descriptor sets in
\code{aid364}:

\begin{Schunk}
\begin{example}
desc.lengths
\end{example}
\begin{Soutput}
#> [1]  24 147
\end{Soutput}
\end{Schunk}

Users can also name the descriptor sets by providing a character vector
to the \code{des.names} argument. If this character vector is specified,
all of \code{ModelTrain} output and downstream \pkg{chemmodlab}
functions will name the descriptor sets accordingly:

\begin{Schunk}
\begin{example}
des.names = c("BurdenNumbers", "Pharmacophores")
\end{example}
\end{Schunk}

The specification of distinct descriptor sets in a data frame is
illustrated in the following call to \code{ModelTrain}:

\begin{Schunk}
\begin{example}
cml <- ModelTrain(d = aid364, ids = TRUE, xcol.lengths = desc.lengths, 
                  des.names = des.names, nfolds = 10, nsplits = 3,
                  seed.in = c(11111, 22222, 33333))
\end{example}
\end{Schunk}

The \code{nsplits} argument sets the number of splits to use for
repeated cross validation and \code{nfolds} sets the number of folds to
use for each cross validation split. The default values have been used.
\code{seed.in} sets the seeds used to randomly assign folds to
observations for each repeated cross-validation split. If NA, the first
seed will be 11111, the second will be 22222, and so on.

If the descriptor set columns are not identified by the user,
\code{ModelTrain} assumes there is one descriptor set, namely all
columns in d except the response column and optional ID column.
Alternatively, the argument \code{xcols} may be used to explicitly
specify the columns corresponding to each descriptor set. Also, if it is
more convenient, descriptor sets can be provided as a list of matrices
with the argument \code{x} and the response as a numeric vector with the
argument \code{y}.

\subsection{chemmodlab models}\label{chemmodlab-models}

Currently, 13 different machine learning models are implemented in
\pkg{chemmodlab}. The details of each modeling method, including
descriptions of the default parameters, are provided at
\url{https://jrash.github.io/chemmodlab/}. Briefly, the current models
are: elastic net (ENet), k-nearest neighbors (KNN), lasso (Lasso), least
angle regression (LAR), neural networks (NNet), partial least squares
linear discriminant analysis (PLSLDA), partial least squares (PLS),
principal components regression (PCR), ridge regression (Ridge), random
forest (RF), two implementations of recursive partitioning (Tree,
RPart), and support vector machines (SVM).

These models have been carefully chosen to provide good coverage of the
spectrum of model flexibility and interpretability available for models
implemented in R. Typically, models that are more flexible (e.g., NNet)
are capable of fitting more complex relationships between predictors and
response, but suffer in terms of model interpretability relative to
other less flexible models (e.g., Lasso). By testing for the significant
differences between model performance measures, we hope to find
interpretable models whose performance is not significantly different
from (i.e., plausibly equivalent to) the best model, suggesting a model
that provides an understanding of the relationship between the
predictors and response that exceeds mere prediction.

Some modeling strategies may not be suitable for both binary and
continous responses. Six of the models have implementations in R that
directly support both binary and continuous responses (Tree, RPart, RF,
KNN, NNet, and SVM). However, six methods (Lasso, LAR, Ridge, ENet, PCR,
and PLS) assume that responses have equal variances and normal
distributions. This assumption is often reasonable for continuous
responses, but may be suspect if the response is binary. For these
latter six methods, binary responses are treated as continuous,
resulting in continuous response predictions that are not restricted to
range between 0 and 1. A threshold can then be applied to obtain a
binary predicted response. The model assessment functions discussed
later allow users to select this threshold. Finally, PLSLDA cannot be
applied to a continous response, but if the user wishes to analyze this
type of data, a threshold value may be used to convert a continuous
response to a binary one.

In cheminformatics applications, descriptors often show strong
multicollinearity. Since this is often problematic for machine learning
models, we have specifically included several models in \pkg{chemmodlab}
that are known to be resilient to multicollinearity (e.g., PCR and PLS).
However, with the exception of PCR, which utilizes uncorrelated linear
combinations of the original descriptors and is a highly interpretable
model, models for which prediction is not considerably affected by
multicollinearity do suffer in terms of model interpretability. For
example, when one variable of a set of highly positively correlated
variables is selected for inclusion in the Lasso linear model, the
selection of the variable is essentially aribitrary. Fitting the same
Lasso model to a slightly different data set would likely result in the
selection of a different variable from the same set.

\pkg{chemmodlab} has been designed in a way that it is easily extensible
to new machine learning modeling methods, and new modeling methods will
be added as the authors identify those that have broad utility to our
users. Support for other models can be requested here:
\url{https://github.com/jrash/chemmodlab/issues}.

\pkg{chemmodlab} automatically performs data preprocessing before
fitting the models that require it (e.g., centering and scaling
variables before PCR), so the user need not worry about preprocessing of
descriptors prior to model fitting.

\subsection{\texorpdfstring{Specifying model parameters with
\code{user.params}}{Specifying model parameters with }}\label{specifying-model-parameters-with}

Sensible default values are selected for each tunable model parameter,
however users may set any parameter manually using the
\code{user.params} argument.

\code{MakeModelDefaults} is a convenience function that makes a list
containing the default parameters for all models implemented in
\code{ModelTrain}. Users may set any parameter manually by generating a
list with this function and modifying the parameters assigned to each
modeling method:

\begin{Schunk}
\begin{example}
params <- MakeModelDefaults(n = nrow(aid364),
 p = ncol(aid364[, -c(1, 2)]), classify = TRUE, nfolds = 10)
params[1:3]
\end{example}
\begin{Soutput}
#> $NNet
#>   size decay
#> 1    2     0
#> 
#> $PCR
#> NULL
#> 
#> $ENet
#>   lambda
#> 1      1
\end{Soutput}
\begin{example}
params$NNet$size <- 10
params[1:3]
\end{example}
\begin{Soutput}
#> $NNet
#>   size decay
#> 1   10     0
#> 
#> $PCR
#> NULL
#> 
#> $ENet
#>   lambda
#> 1      1
\end{Soutput}
\end{Schunk}

This list can then be provided to the \code{user.params} argument to
assign the tuning parameter values used by \code{ModelTrain}:

\begin{Schunk}
\begin{example}
cml <- ModelTrain(USArrests, models = "NNet", nsplits = 3,
 user.params = params)
\end{example}
\end{Schunk}

\section{Model assessment}\label{model-assessment}

\subsection{\texorpdfstring{Repeated \(k\)-fold
cross-validation}{Repeated k-fold cross-validation}}\label{repeated-k-fold-cross-validation}

For each descriptor set, \code{ModelTrain} performs repeated \(k\)-fold
cross validation for the selected set of regression and/or
classification models.

For each cross-validation split, observations are randomly assigned to
one of \(k\) folds, splitting the data set into \(k\) blocks that are
approximately equal in size. The number of cross validation folds
(\(k\)) is set with the \code{nfolds} argument. Users may also use the
\code{seed.in} argument to set the seed for each split, so that the
\code{ModelTrain} results are reproducible. Each block is iteratively
held out as a test set, while the remaining \(k-1\) blocks are used to
train each D-M combination. Predictions for the held out test set are
then made with the resulting models.

Many resampling methods for assessing model performance involve
partitioning a data set into a training set and test set. With these
methods, predictions are made on observations that are not involved in
training. This results in model performance measures that are less
likely to reward over-fitting.

Since performance measures can be highly variable \citep{James2013}
depending on which observations are held out and which are involved in
training, the repetition of this procedure during \(k\)-fold cross
validation and the averaging of the performance measures often result in
a more accurate estimation of model performance than a one-time split.

Finding the right number of cross-validation folds for the estimation of
a performance measure involves consideration of the
\textbf{bias-variance trade off}. The mean squared error of an
estimator, a measure of how accurately an estimator estimates the true
value of a parameter, can be partitioned into two components, bias and
variance:

\[E[(\hat{\theta} - \theta)^2] = (E[\hat{\theta}] - \theta)^2 + Var[\hat{\theta}],\]

\noindent where \(\hat{\theta}\) is the estimator of the true
performance measure, \(\theta\), for the population of test sets similar
to the data set under consideration. The first component is squared bias
and the second is variance. An increase in either the bias or variance
will decrease the quality of an estimator. When a resampling method
substantially over- or under-estimates a performance measure on average,
it is said to have high \textbf{bias}. Bias is often related to size of
the data set that is held out as a test set \citep{James2013}. The
smaller the number of folds in \(k\)-fold cross validation, the more
observations are held out in each fold, and the less observations that
are used to train a model. Fewer observations in a training set means
that a model is likely to perform worse, and model predictions tend to
miss the target. Thus, performing \(k\)-fold cross validation with 2
folds, where there is 50\% of the data in each fold, would likely result
in high bias.

In contrast, a performance measure estimator suffers from high
\emph{variance} when its estimate varies considerably when there are
slight changes made to the training and/or test set.
Leave-One-Out-Cross-Validation (LOOCV) refers to \(k\)-fold cross
validation with \(k\) equal to \(n\), the number of observations. LOOCV
often suffers from high variance \citep{James2013}. This is due to the
fact that the training set changes very little with each iteration of
LOOCV. Thus, performance measure estimates tend be highly positively
correlated. The mean of a highly correlated variable has higher variance
than an uncorrelated one. Decreasing the number of folds tends to
decrease the correlation of performance measure estimates and lower the
variance. Therefore, the ideal number of folds to use for cross
validation is often somewhere between 2 and \(n\). The number of folds
often used for \(k\)-fold cross validation are 5 and 10, as these values
frequently provide a good balance between bias and variance.

Several studies \citep{Molinaro2005, Kim2009, Shen2011} have shown that
repeated cross validation can reduce the variance for a \(k\)-fold cross
validation procedure with low bias, achieving a more accurate estimation
of model performance. When \(k\)-fold cross validation is repeated in
\pkg{chemmodlab}, multiple iterations of random fold assignment, or
``splits'', are performed. Because the observed performance measures may
vary greatly depending on the definition of folds during \(k\)-fold
cross validation, all models are built using the same fold assignment,
thus using fold definition as a ``blocking factor'' in the assesment
investigation. This process is repeated for multiple \(k\)-fold cross
validation runs. The user may choose the number of these splits with the
\code{nsplit} argument.

There are a myriad of potential model violations that may lead to over-
or under-estimation of model accuracy measures. These include but are
not limited to: a misspecification of the structural form of the model
(e.g., more descriptors, or a function of the descriptors should be
used), correlation between responses, measurement error in predictors,
non-constant variance in the response, collinearity between descriptors,
and non-normality of the response. Also, the types of model violations
that occur in the training and test sets generated by \(k\)-fold cross
validation may vary depending on how the data is split, resulting in
over- or under-estimation of a performance measure depending on how the
data was split. Since it is implausible to account for all model
violations, our approach is to treat the effect of model violation as a
nuisance variable. By doing several random splits and averaging the
\(k\)-fold cross validated performance measures, we ``average out'' the
model violation effect on the performance measure estimate.

We will see later how, in addition to increased accuracy in estimation
of performance measures, repeated cross validation allows one to measure
the standard error of model performance measures, quantifying how
sensitive performance measures are to fold assignments.

\subsection{Accumulation curves}\label{accumulation-curves}

\code{plot.chemmodlab} takes a \pkg{chemmodlab} object output by the
\code{ModelTrain} function and creates a series of accumulation curve
plots for assessing model performance.

The accumulation curve for binary responses shows the number of
positives versus the number of ``tests'' performed, where testing order
is determined by the \(k\)-fold cross validated predicted probability of
a response being positive. The \code{max.select} argument sets the
maximum number of tests to plot for the accumulation curves. By default,
\(\left \lfloor{{\min(300, \frac{n}{4})}}\right \rfloor\) is used, where
\(n\) is the number of observations. This prioritizes finding actives in
a relatively small number of tests.

Two series of plots are constructed. In the ``methods'' series, there is
one plot per CV split and descriptor set combination. The accumulation
curves for each modeling method are plotted together so that they can be
compared. In the ``descriptors'' plot series, there is one plot per CV
split and model fit. The accumulation curves for each descriptor set are
plotted together so that they can be compared.

By default a large number of accumulation curve plots are constructed.
The \code{splits} and \code{meths} arguments may be used to only plot a
subset of splits in the ``methods'' series of plots and methods in the
``descriptors'' series of plots, respectively. The \code{series}
argument specifies whether the ``methods'' series of plots, the
``descriptors'' series of plots, or both are generated.

\begin{Schunk}
\begin{example}
plot(cml, splits = 1, series = "descriptors")
\end{example}
\begin{figure}

\includegraphics{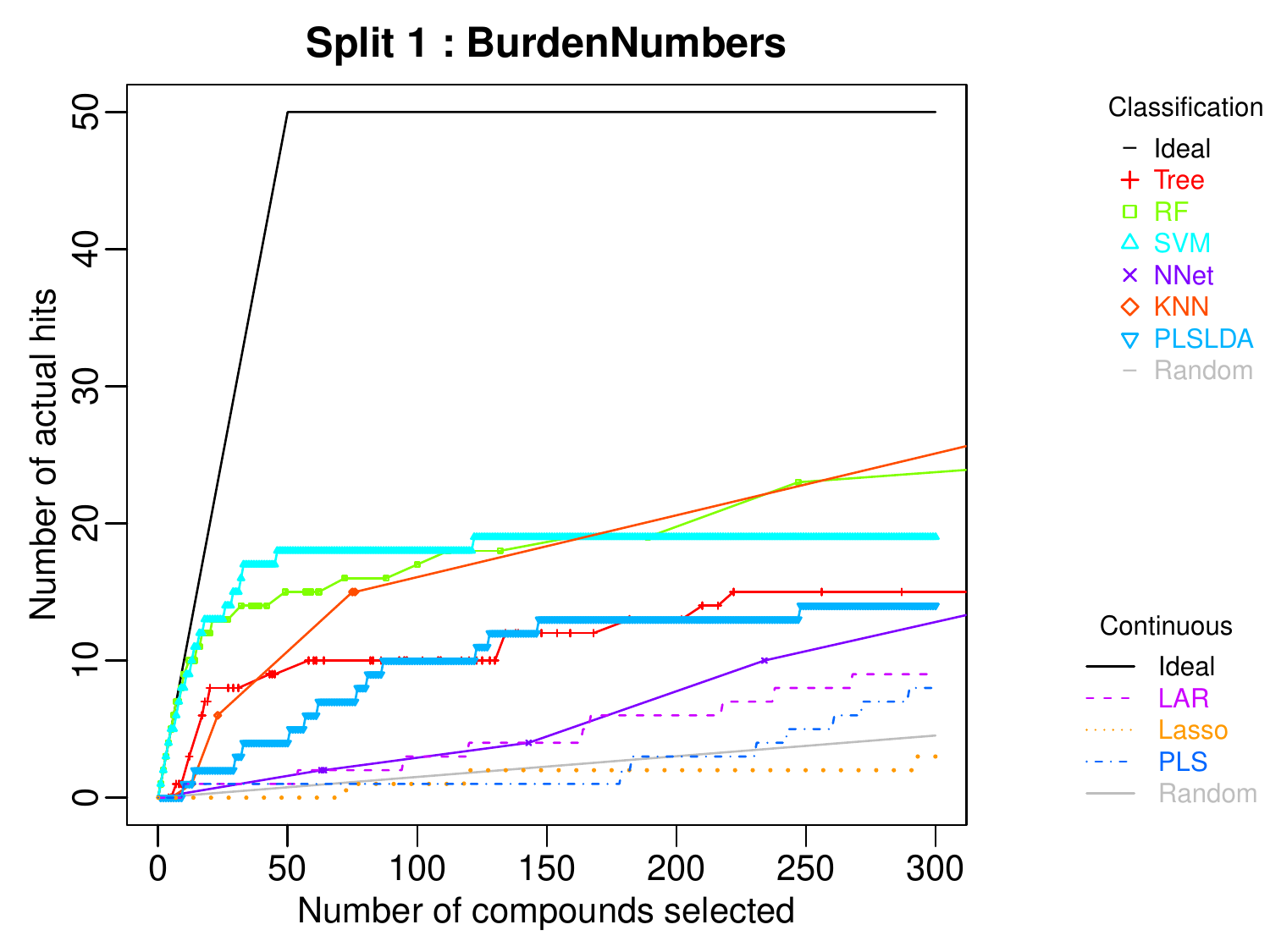} \includegraphics{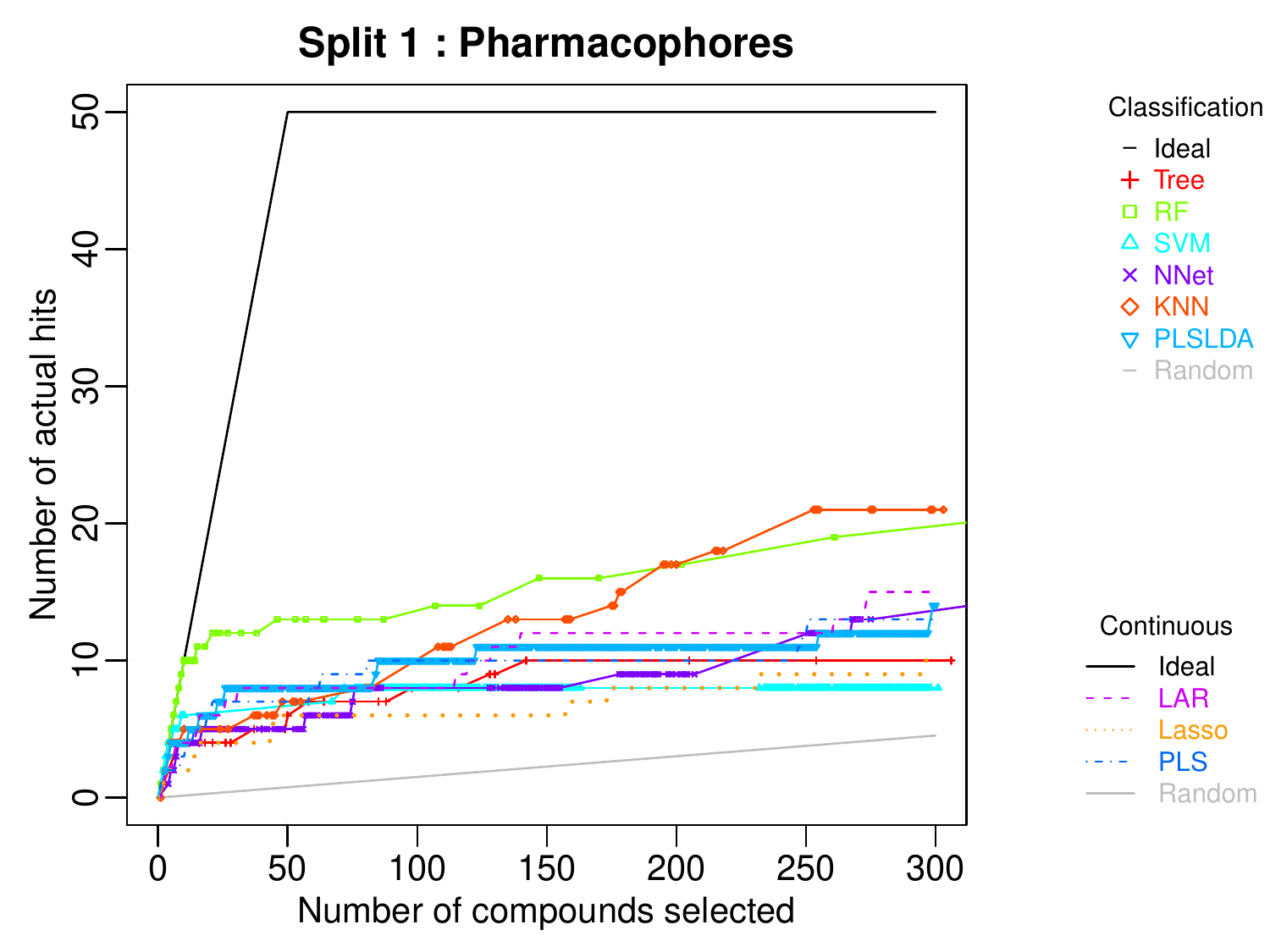} \hfill{}

\caption{\label{fig:acc_curves1}Comparison of the first CV split accumulation curves for models fit to the Burden Numbers descriptor set (top) and the Pharmacophore descriptor set (bottom)}\label{fig:acc_curves1}
\end{figure}
\end{Schunk}

\begin{Schunk}
\begin{example}
plot(cml, splits = 1, meths = c("SVM", "RF"), series = "methods")
\end{example}
\begin{figure}

\includegraphics{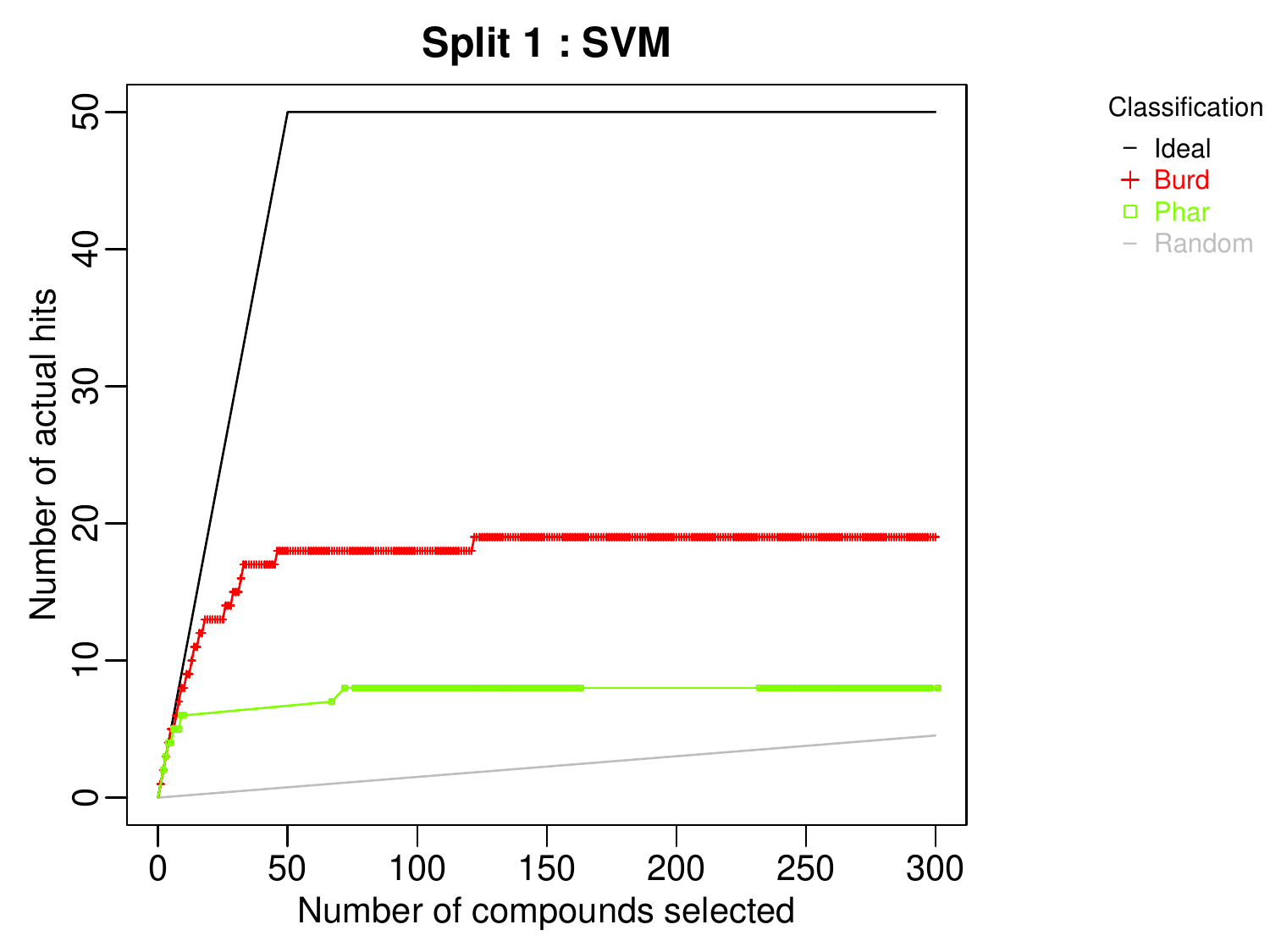} \includegraphics{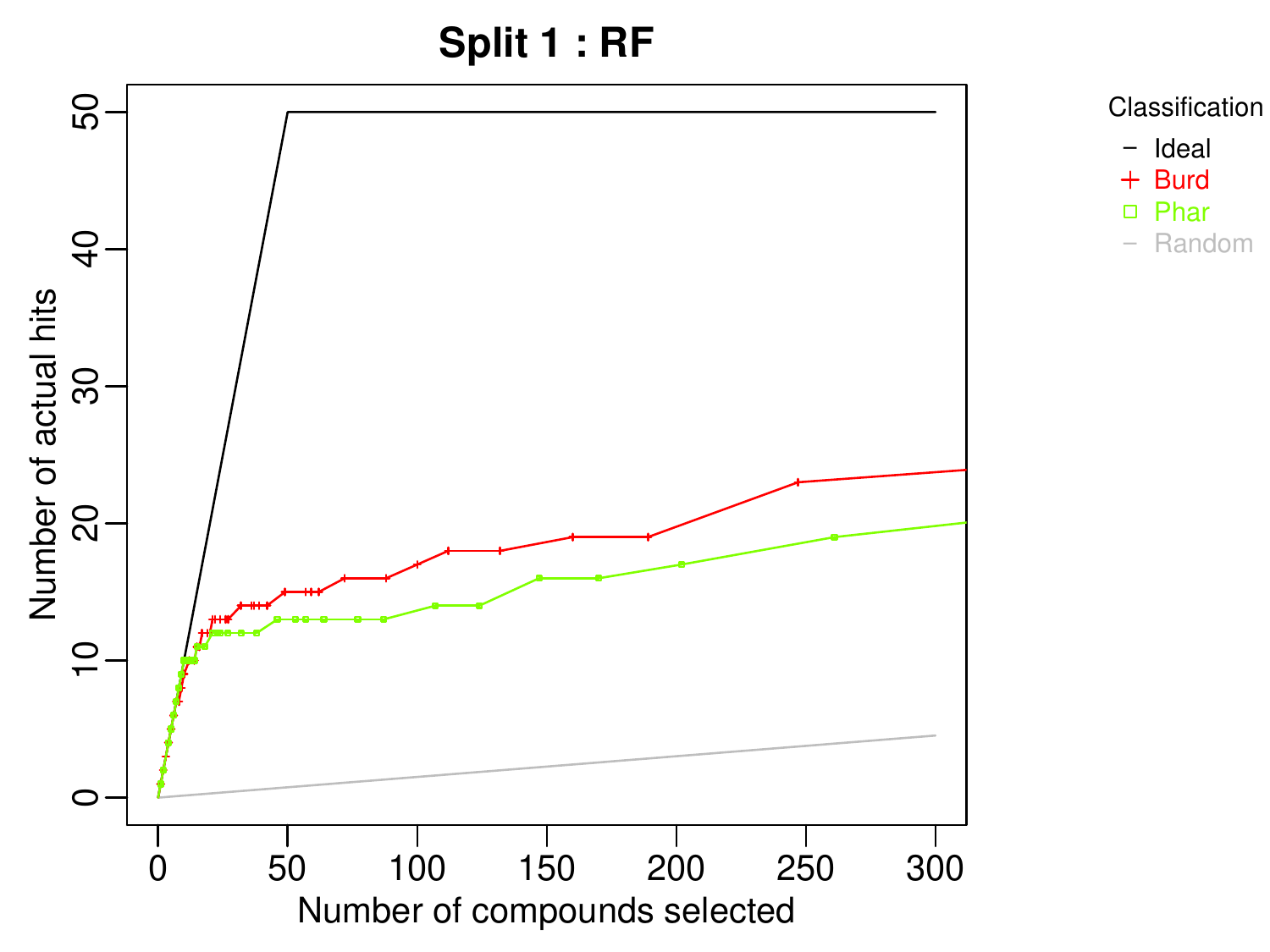} \hfill{}

\caption{\label{fig:acc_curves2}Comparison of the first CV split descriptor set accumulation curves for the SVM method (top) and the RF method (bottom)}\label{fig:acc_curves2}
\end{figure}
\end{Schunk}

An ``ideal'' curve is plotted on these graphs to demonstrate the
accumulation curve of a model that correctly identifies the \(p\)
positives in the first \(p\) tests. Thus, at \(m\) tests, models with
accumulation curves that are nearest to the ideal curve are preferable.
Also, if an accumulation curve has a slope that is parallel to the ideal
curve for an interval of tests, the model has ideal performance for that
interval. A ``random'' curve shows the accumulation curve if the testing
order were decided at random. At \(m\) tests, models with accumulation
curves that are below the random curve have worse performance than
random chance. Models that were fit as classification models (see
``Model fitting: chemmodlab models'') are represented as solid lines
with different colors and shapes specifying the modeling method. Models
that were fit as continuous models and then thresholded are
differentiated by line type and color.

In Figure \ref{fig:acc_curves1}, we have plotted the accumulation curves
for the first CV split, generating two plots in the ``methods'' series,
one for each descriptor set. Comparing the accumulation curves for
models utilizing Burden Number descriptors, SVM and RF have much better
hit rates than the other models for the initial 100 compounds
prioritized for testing. However, if more than 100 tests were to be
performed, the KNN method would have the best performance. Intersections
between model accumulation curves indicate the number of tests at which
one model's performance overtakes another's. Considering the
Pharmacophore fingerprints, the RF method has ideal performance
initially, but is eventually superceded by the KNN method, as was the
case for the Burden Number descriptors.

The next series of plots (Figure \ref{fig:acc_curves2}) is the
``methods'' series. When the Split 1 descriptor set performances are
compared for the SVM method, the Pharmacophore fingerprints have
considerably worse performance than the Burden Numbers. This is sensible
as SVM performance often suffers in high dimensional spaces. Comparing
descriptor set performances for the RF method, the Burden Number
descriptors have slightly improved performance over the Pharmacophore
fingerprints for all but the initial few tests. However, at any number
of tests at which Burden Numbers outperform Pharmacophores, Burden
Numbers only provide a few more hits. Therefore, it is plausible that if
this experiment were performed again with a different, but similar data
set, Pharmacophores would perform equally as well as Burden Numbers. One
may be concerned about the statistical significance of this improved
performance. This observation has motivated the construction of the
\code{CombineSplits} function, which rigorously tests for statistically
significant differences.

The accumulation curve has also been extended to continuous responses.
In QSAR models, a continuous response is often a measure of binding
affinity (e.g., pKi) where a large positive value is preferable.
Therefore, in these accumulation curves, testing order is determined by
ordering the predicted response in decreasing order. The response is
then accumulated so that \(\sum_{i=1}^{m} y_i\) is the sum of the \(y\)
over the first \(m\) tests. The binary response accumulation curve is a
special case of this.

\subsection{Multiple comparisons similarity
plot}\label{multiple-comparisons-similarity-plot}

\code{CombineSplits} evaluates a specified performance measure across
all splits. This function assesses how sensitive performance measures
are to fold assignments, or small changes to the training and test sets.
Intuitively, this assesses how much a performance measure may vary if
predictions were made on a test set that is similar to the data set
analyzed. Multiplicity-adjusted statistical tests are used to determine
the best performing D-M combination.

As input, \code{CombineSplits} takes a \pkg{chemmodlab} object produced
by the \code{ModelTrain} function. \code{CombineSplits} can use many
different performance measures to evaluate binary classification model
performance (namely: error rate, sensitivity, specificity, area under
the receiver operating characteristic curve, positive predictive value
also known as precision, F1 measure, and initial enhancement). By
default, \code{CombineSplits} uses initial enhancement proposed by
\citet{Kearsley} to assess model performance. Initial enhancement at
\(m\) tests is the hit rate -- the fraction of accumulated positives at
\(m\) tests -- divided by the proportion of positives in the entire data
set. This is a measure of a model's hit rate fold improvement over
random chance. A desirable model will have an initial enhancement much
larger than one. A typical number of tests for initial enhancement is
\(m=300\).

\begin{Schunk}
\begin{example}
CombineSplits(cml)
\end{example}
\begin{Soutput}
#>    Analysis of Variance on: 'enhancement'
#>  Using factors: Split and Descriptor/Method combination
#> Source    DF        SS        MS         F   p-value   
#> Model     19   72.5318    3.8175   22.8131    <.0001   
#> Error     34    5.6894    0.1673   
#> Total     53   78.2212   
#>       R-Square   Coef Var   Root MSE       Mean   
#>         0.9273    12.9890     0.4091     3.1494   
#> Source       DF       SS       MS        F   p-value   
#> Split         2    1.463    0.732    4.372    0.0194   
#> Desc/Meth    17   71.069    4.181   24.983    <.0001
\end{Soutput}
\begin{figure}

\includegraphics{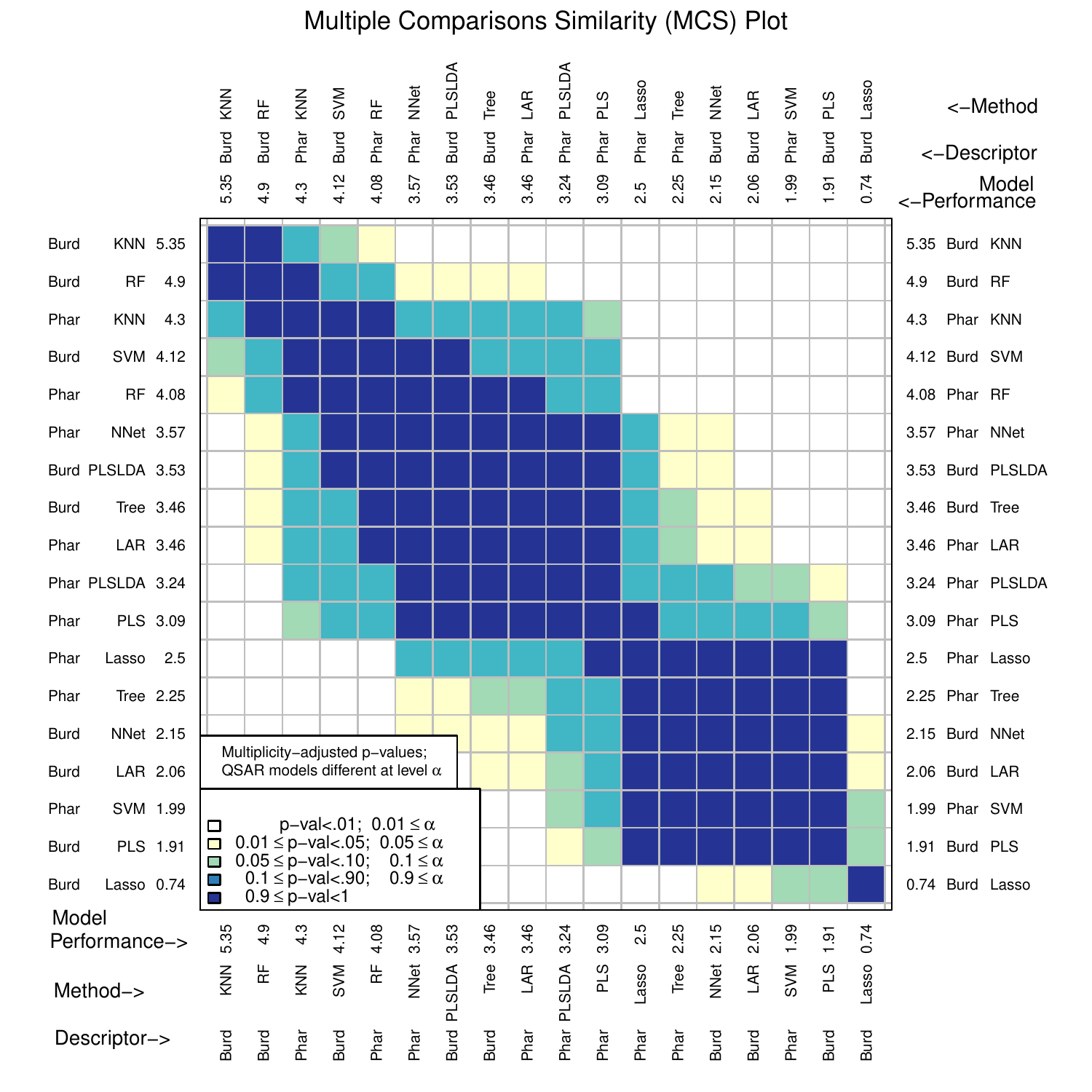} \hfill{}

\caption{\label{fig:CombineSplits_ie}MCS plot using initial enhancement at 300 tests as the performance measure.}\label{fig:CombineSplits_ie}
\end{figure}
\end{Schunk}

An advantage of performing repeated k-fold cross validation with
\code{ModelTrain} is that the output can be viewed as a designed
experiment with two factors: D-M combination and split (fold
assignment). Therefore, \code{CombineSplits} performs an analysis of
variance (ANOVA) to identify significant differences between the mean
performance measures according to factor levels. The linear model
corresponding to this ANOVA is:

\[Y_{ijk} = \mu + \alpha_i + \beta_j + \epsilon_{ijk},\]

\noindent where \(\alpha_i\) corresponds to ith level of the split
factor and \(\beta_j\) to the jth level of the D-M combination factor.
From the ANOVA table in this example, the split main effect is
marginally significant (p-value of 0.0194), indicating that there is a
significant difference between mean initial enhancement across splits,
averaging over D-M combinations. In other words, some splits result in
significantly larger initial enhancement values than other splits. This
endorses our decision to treat splits as a blocking factor. The D-M
combination main effect is highly significant (p-value \textless{}
0.0001). The ``Error MS'' estimates the variance in the performance
measures within the groups corresponding to each combination of factor
levels. The ``Model MS'' estimates the variance between groups.

The multiple comparisons similarity plot in Figure
\ref{fig:CombineSplits_ie} shows the results for tests of significance
among all pairwise differences of mean model performance measures. Along
both the x- and y-axes, D-M combinations are ordered from best to worst
performance measure. Because there can potentially be a large number of
methods (18 in the example leading to \({18 \choose 2} = 153\) pairwise
comparisons), an adjustment for multiple testing is necessary. We use
the Tukey-Kramer multiple comparison procedure (see \citet{Tukey1994}
and \citet{Kramer1956}).

In Figure \ref{fig:CombineSplits_ie}, Burden Numbers-KNN, Burden
Numbers-RF, Pharmacophore-KNN, and Burden Numbers-SVM are the top
performing models. We conclude that for the population of compounds that
are similar to the data set analyzed, the initial enhancement of these
four models is plausibly the same. The intuition is that if predictions
were made on a new test set that is similar to the data set under
consideration, these four models would plausibly have very similar
prediction performance.

There are different characteristics of the top performing models that
may lead to a researcher choosing one over the other. While KNN is not
the most interpretable model, there are fast heuristic KNN regression
methods implemented in R such as
\href{https://cran.r-project.org/web/packages/FNN/index.html}{FNN}
\citep{Beygelzimer2013} that make predicting on large data sets more
manageable. However RF has a time complexity comparable to the fast
heuristic KNN, and also results in a more interpretable model. Measures
of variable importance can be computed for RF, which allow users to
identify the subset of variables that are most important for prediction.
The Burden Numbers-SVM model is the least interpretable of the set and
has time complexity that scales the worse with \(n\) (between \(O(n^2)\)
and \(O(n^3)\) time complexity). Though the Burden Numbers-KNN model is
the best performing model according to mean initial enhancement, this
model performance measure is only .45 larger than the Burden Numbers-RF
model and the difference is not statistically significant. Since the
Burden Numbers-RF model is more interpretable and works just as well
with large data, this modeling method may be preferrable.

The Pharmacophore-RF model is only significantly different from the best
performing model at a level between .01 and .05, but may be preferrable,
as both the descriptor set and the model are highly interpretable.

For many applications, users may know the number of tests they would
like to perform. This is often the case in drug discovery when chemists
have a set number of compounds they would like to assay and the goal is
to enrich this set of compounds with as many actives as possible. The
number of tests used for initial enhancement may be modified with the
\code{m} argument (Figure \ref{fig:CombineSplits_ie_100}):

\begin{Schunk}
\begin{example}
CombineSplits(cml, m = 100)
\end{example}
\begin{figure}

\includegraphics{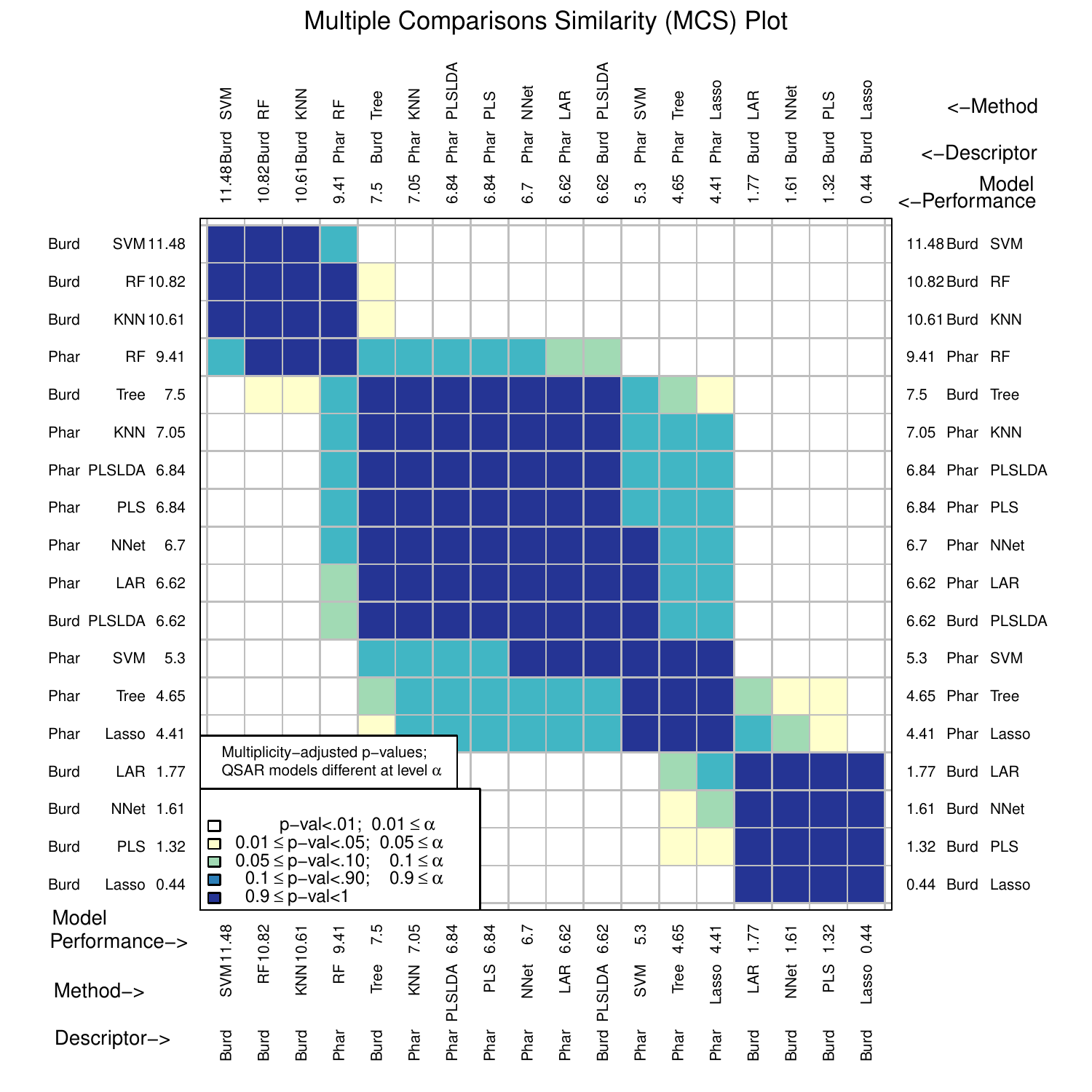} \hfill{}

\caption{\label{fig:CombineSplits_ie_100}MCS plot using initial enhancement at 100 tests as the performance measure.}\label{fig:CombineSplits_ie_100}
\end{figure}
\end{Schunk}

Figure \ref{fig:CombineSplits_ie_100} underscores the practical
importance of using an appropriate number of tests for initial
enhancement. If a significance level of .05 were to be used, the
Pharmacophore-KNN model is no longer among the best performing models at
\(m=100\), while the highly interpretable model, Pharmacophore-RF, is
now among the best. This model is the clear choice if an interpretable
model is desired.

These results support the observations regarding descriptor set
accumulations curves for the RF model in Figure \ref{fig:acc_curves2}.
While there appeared to be a slight improvement using Burden Number
descriptors in lieu of Pharmacophore fingerprints, this improvement may
not be considered statistically significant.

In this particular example, specificity does not do a good job at
distinguishing the best performing models, as the set of plausibly best
performing models is quite large (Figure \ref{fig:CombineSplits_sp}).
This is due to the fact that there are many models that have an average
specificity that is similar to the best performing model:

\begin{Schunk}
\begin{example}
CombineSplits(cml, metric = "specificity")
\end{example}
\begin{figure}

\includegraphics{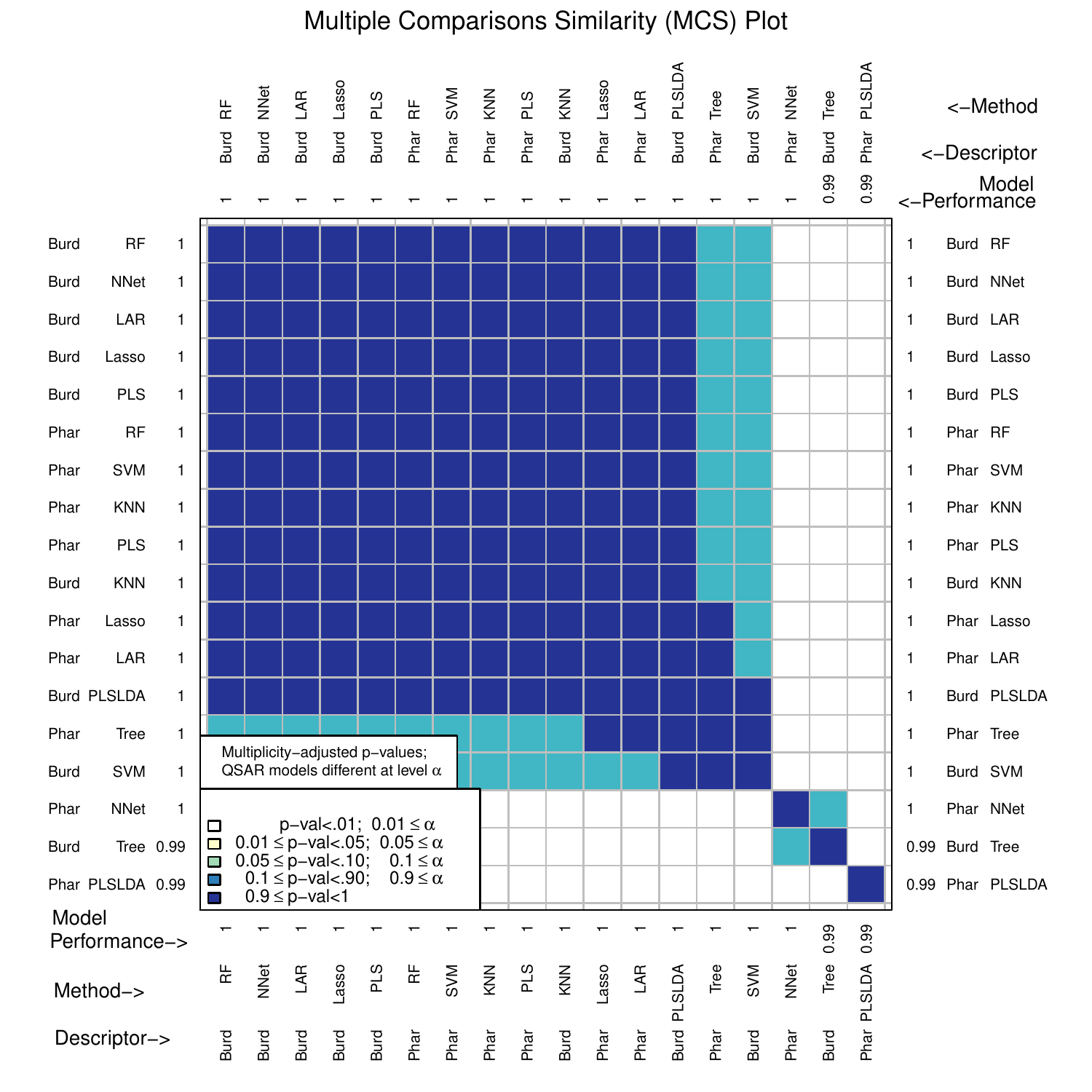} \hfill{}

\caption{\label{fig:CombineSplits_sp}MCS plot using specificity for all compounds as the performance measure.}\label{fig:CombineSplits_sp}
\end{figure}
\end{Schunk}

Sensitivity, however, distinguishes the best performing model much
better (Figure \ref{fig:CombineSplits_se}):

\begin{Schunk}
\begin{example}
CombineSplits(cml, metric = "sensitivity")
\end{example}
\begin{figure}

\includegraphics{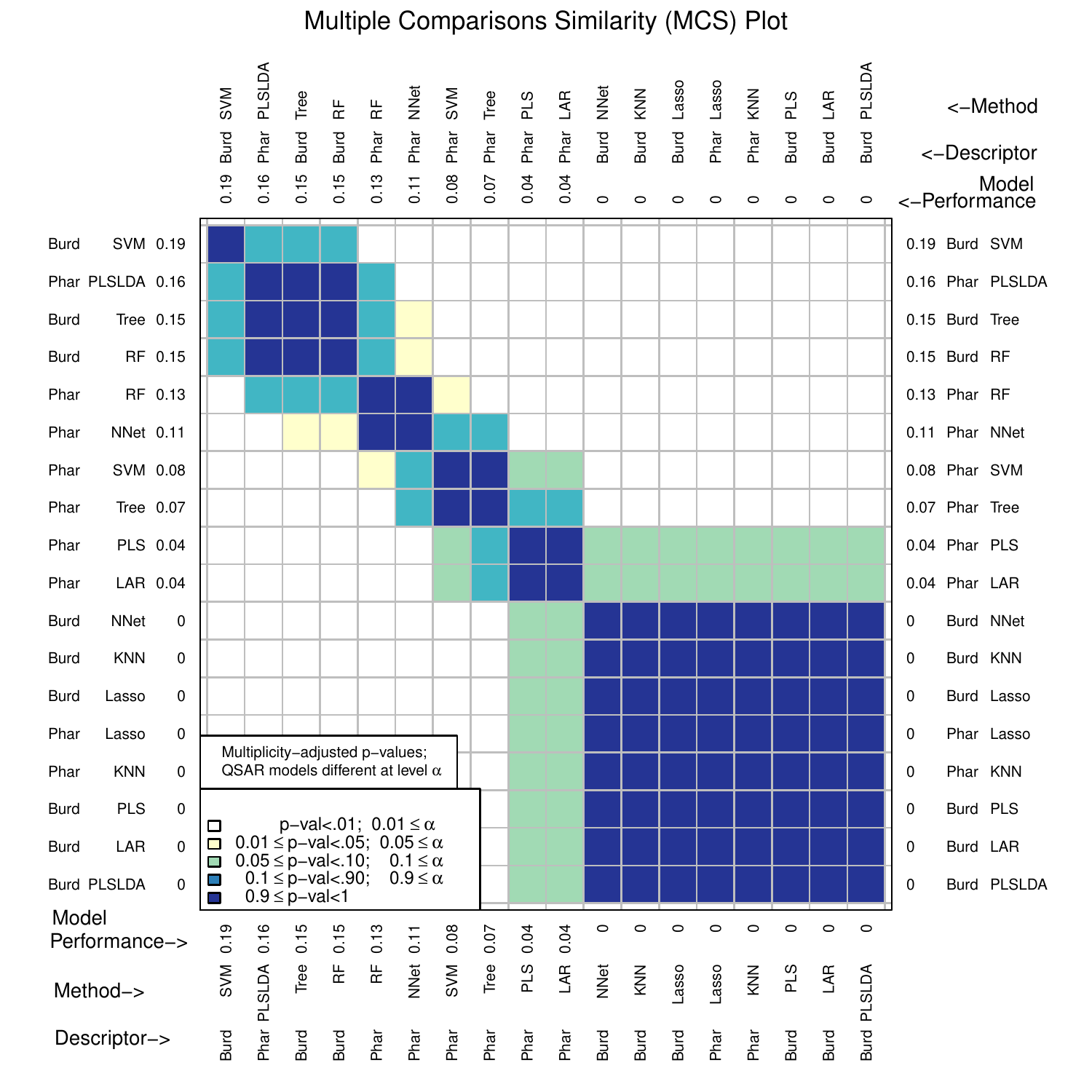} \hfill{}

\caption{\label{fig:CombineSplits_se}MCS plot using sensitivity for all compounds as the performance measure.}\label{fig:CombineSplits_se}
\end{figure}
\end{Schunk}

For binary responses, model performance may also be assessed with
misclassification rate (Figure \ref{fig:CombineSplits_er}). However,
this measure may be inappropriate for drug discovery because it equally
penalizes false positives and false negatives. As we will explain in
detail later on, researchers in cheminformatics tend to be less
concerned with false negatives and instead prioritize finding actives
early. Therefore, a meaure like intial enhanecement may be more
appropriate.

\begin{Schunk}
\begin{example}
CombineSplits(cml, metric = "error rate")
\end{example}
\begin{figure}

\includegraphics{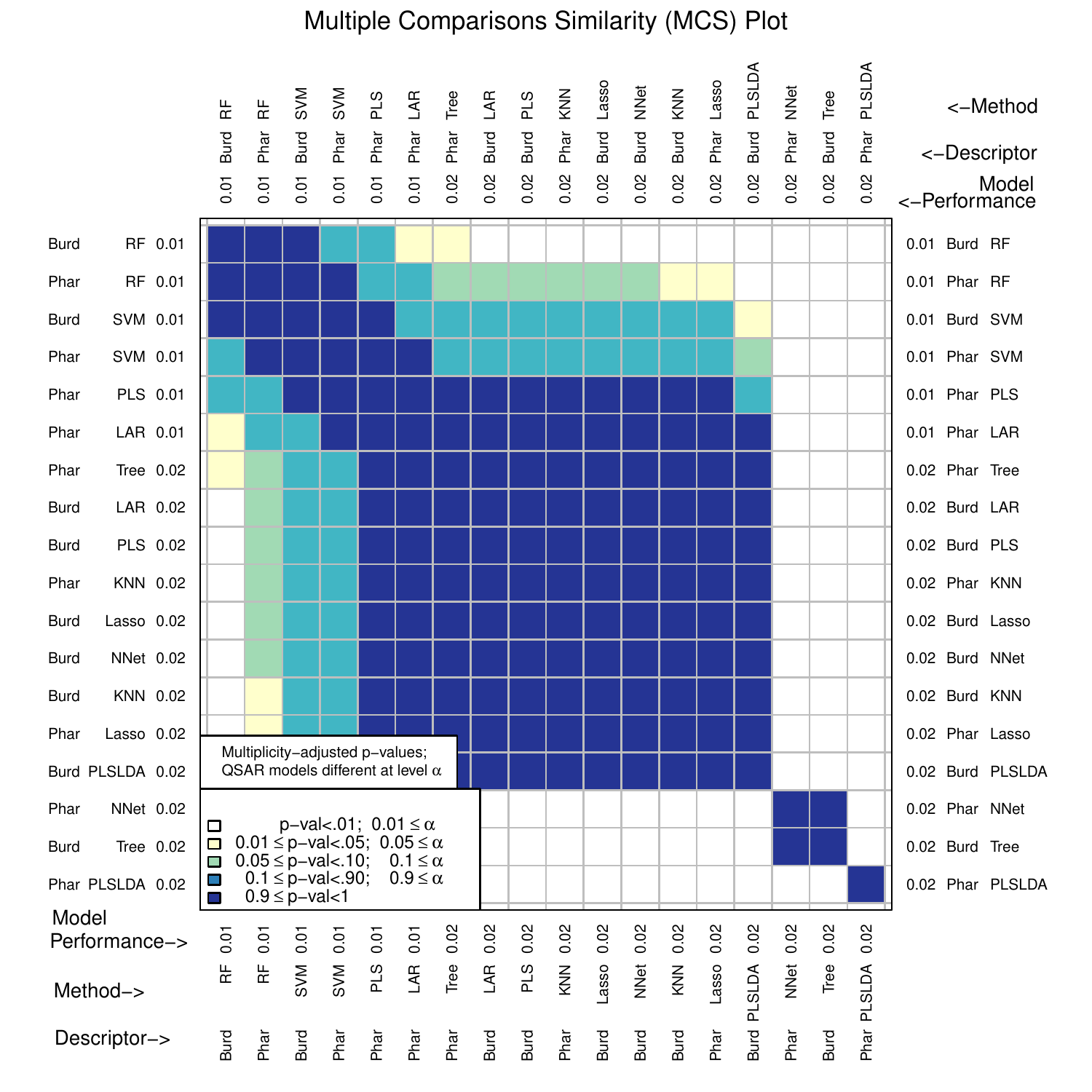} \hfill{}

\caption{\label{fig:CombineSplits_er}MCS plot using error rate for all compounds as the performance measure.}\label{fig:CombineSplits_er}
\end{figure}
\end{Schunk}

This example illustrates how using misclassification rate can be
misleading if this is the only model performance measure.
Misclassification rate suggests several models (such as
Pharmacophore-PLS and Pharmacophore-LAR) are among the best performing
models, but these models actually have very low sensitivity (both .04).
These models identify only a few of the 50 true positives in the data
set. If the goal of a medicinal chemist is to identify all of the active
compounds in their data set, these models may actually have poor
performance.

It is also possible that models with high sensitivity incorrectly
identify many compounds as active. The positive predictive value (PPV),
also known as precision, measures the percentage of compounds that were
correctly predicted to be active. PPV is an important means of
evaluating search engines. A search engine often finds a multitude of
potentially relevant documents, but a user is only capable of looking at
the first few results. Search engines need to have very high PPV, even
at the expense of sensitivity and specificity \citep{Brin1998}. The goal
is not to correctly identify all the relevant documents or all the
irrelevant ones, but to identify the most relevant documents in the top
few results returned. In the context of drug discovery, testing
potential drugs can be expensive and time consuming. Medicinal chemists
can only test a small proportion of the compounds in their data set, so
the goal is to test compounds only when the certainty of activity is
high. Therefore, models with low PPV (i.e., a high false positive rate)
may be less than ideal.

Figure \ref{fig:CombineSplits_ppv} shows that the best performing models
according to PPV (Pharmacophore-SVM, Pharmacophore-RF,
Pharmacophore-PLS, Burden Numbers-RF) all have a perfect PPV value --
every compound predicted to be active was indeed active.

\begin{Schunk}
\begin{example}
CombineSplits(cml, metric = "ppv")
\end{example}
\begin{figure}

\includegraphics{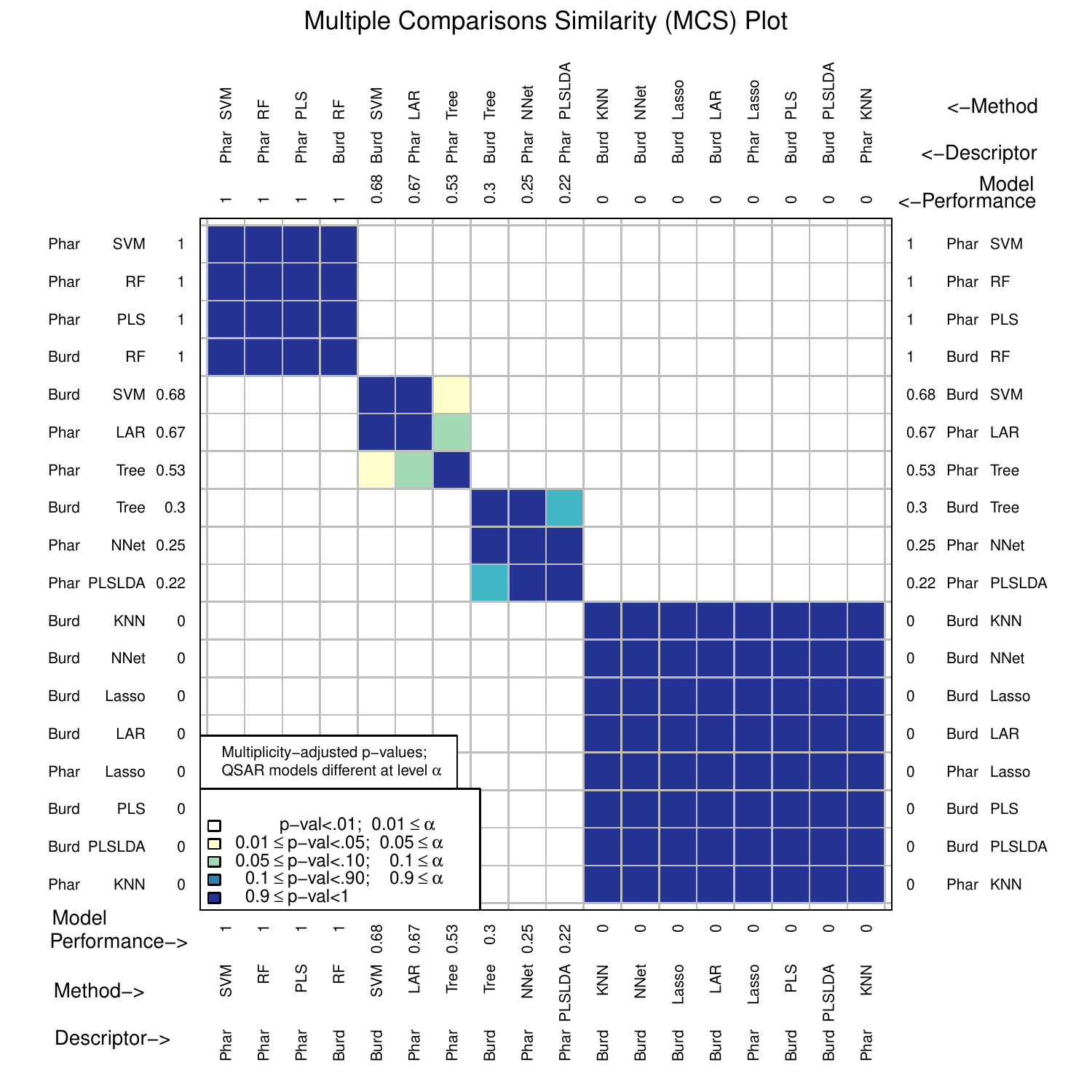} \hfill{}

\caption{\label{fig:CombineSplits_ppv}MCS plot using PPV (precision) for all compounds as the performance measure.}\label{fig:CombineSplits_ppv}
\end{figure}
\end{Schunk}

When a model has both high sensitivity and PPV, this means that many of
the actives in the data set were identified with a low number of false
positives. The F1 measure strives to strike this balance between
sensitivity and PPV. It is the harmonic mean of sensitivity and PPV.
Burden Numbers-SVM and Burden Numbers-RF are the models which find the
best balance, with Pharmacophore-RF being only marginally significantly
different (Figure \ref{fig:CombineSplits_fmeasure}).

\begin{Schunk}
\begin{example}
CombineSplits(cml, metric = "fmeasure")
\end{example}
\begin{figure}

\includegraphics{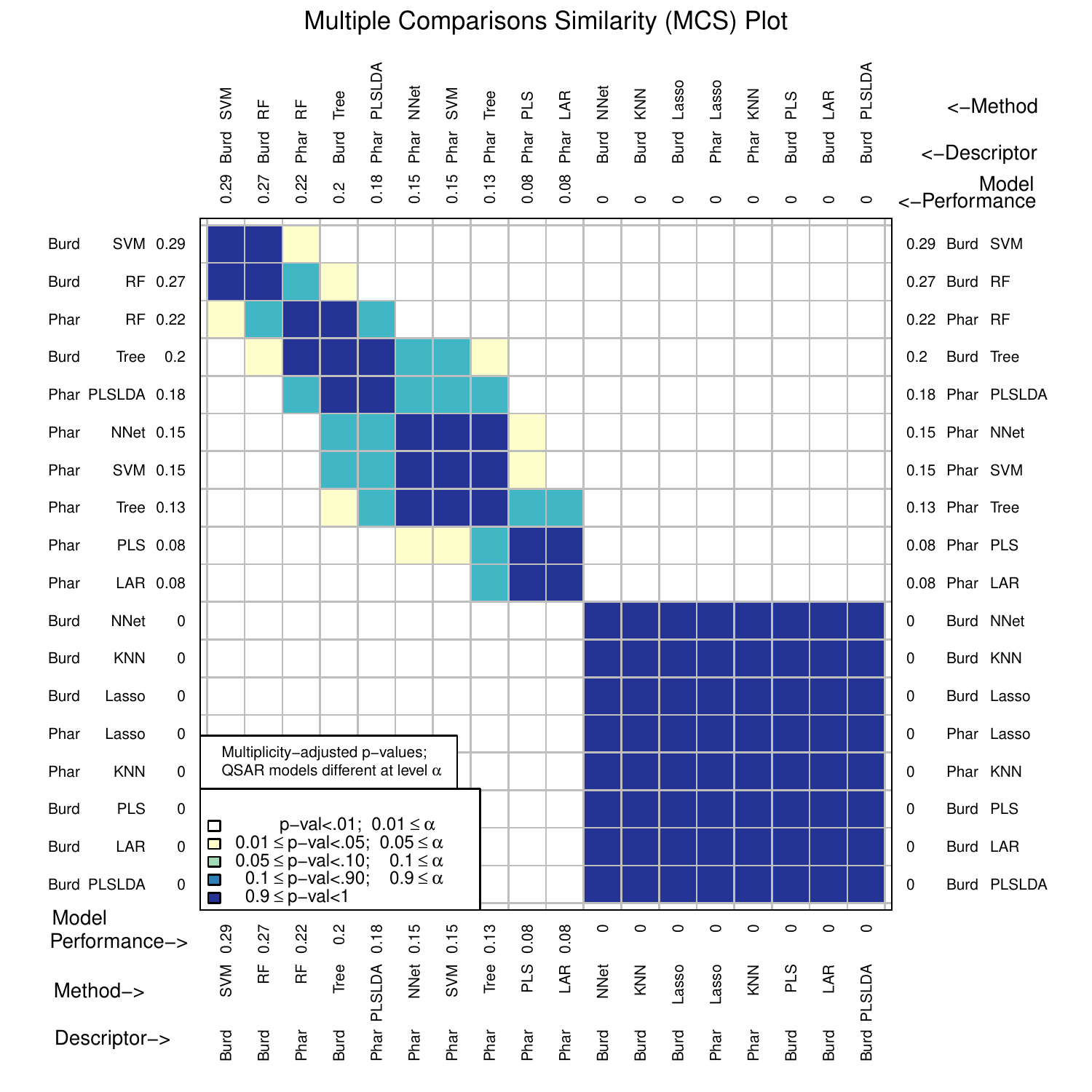} \hfill{}

\caption{\label{fig:CombineSplits_fmeasure}MCS plot using F1 measure for all compounds as the performance measure.}\label{fig:CombineSplits_fmeasure}
\end{figure}
\end{Schunk}

The area under the receiver operating characteristic curve (AUC) has
also been implemented in \pkg{chemmodlab} (Figure
\ref{fig:CombineSplits_auc}).

Table \ref{tab:checks} summarizes the performance of the top performing
D-M combinations over all performance measures considered. Burden
Numbers-RF and Burden Numbers-SVM are consistently the top performers.
However, if a significance level of .01 were used, Pharmacophore-RF
would be a top performer across all performance measures. The fact that
Pharmacophore-RF is among the best models for sensitivity and PPV but is
marginally significantly different for AUC suggests that there may be a
threshold for the predicted probabilities other than .5 that will lead
to better prediction performance for Burden Numbers-RF and Burden
Numbers-SVM. Conversely, the fact that Burden Numbers-SVM is among the
best for AUC but not for sensitivity or PPV suggests that a threshold
other than .5 should be used.

\begin{table}[ht]
\centering
\begin{tabular}{ccccccccccc}
  \toprule
 Descriptor & Model & IE 300 & IE 100 & Spec & Sens & Error Rate & PPV & F1 & AUC \\ 
  \midrule
Burd & RF & \checkmark & \checkmark & \checkmark & \checkmark & \checkmark & \checkmark & \checkmark & \checkmark \\ 
Burd & SVM & \checkmark & \checkmark & \checkmark & \checkmark & \checkmark &  & \checkmark & \checkmark \\ 
Phar & RF & \checkmark - & \checkmark & \checkmark & \checkmark & \checkmark & \checkmark & \checkmark - & \checkmark - \\ 
Burd & KNN & \checkmark & \checkmark & \checkmark &  &  & \checkmark &  & \checkmark \\ 
Phar & SVM &  &  & \checkmark &  & \checkmark & \checkmark &  &  \\ 
   \bottomrule
\end{tabular}
\caption {The best performing D-M combinations according to all performance measures considered.  Check indicates the D-M combination was among the best performers according to a performance measure using a significance level of 0.05.  Check minus indicates marginal significant difference between the D-M combination and the best performer (significance level between .01 and .05). Performance measures considered were: initial enhancement at 300 tests, intial enhancement at 100 tests, specificity, sensitivity, error rate, posterior predictive value, F1 measure, and area under the receiver operating characteristic curve.} \label{tab:checks}
\end{table}

Several performance measures have been included for continuous
responses. Though root mean squared error (RMSE) is used broadly in
statistics, it may not be suitable for continuous chemical assay
responses used in cheminformatics. This is because under-predicting and
over-predicting biological activity is equally penalized. An appropriate
alternative may be initial enhancement. Other options are the
coefficient of determination (\(R^2\)) and Spearman's \(\rho\).

\section{Conclusions and future
directions}\label{conclusions-and-future-directions}

\pkg{chemmodlab} provides a comprehensive collection of methods for
fitting and assessing machine learning models. While these methods have
been selected for their utility to the cheminformatics community, they
can be applied to any data set with binary or continuous variables.
These methods are applicable to a wide range of research areas, and some
model assessment approaches (the MCS plot, continuous valued
accumulation curves) are novel additions to the assessment of the
machine learning methods available in R. The functions in
\pkg{chemmodlab} aim to enable researchers to try many different model
fitting and assessment procedures with ease.

\pkg{chemmodlab} has many future directions in store. Parallel
processing will be utilized so that model fitting for different
descriptor sets, splits, and cross validation folds can be done in
parallel. We also plan to interface \pkg{chemmodlab} with
\textbf{caret}, employing their functions for model tuning during our
model fitting procedure. Support for categorical variables with more
than two levels will be arriving soon. We will also incoporate more
extensions to the accumulation curve as this approach is used
extensively in drug discovery. Extensions will include the area under
the accumulation curve as an assessment measure and the construction of
a mean accumulation curve over multiple splits. Error bars can be
plotted for these curves so that significant differences between D-M
combinations can be analyzed across the entire curves. Additional
graphical output is planned, e.g., to provide receiver operating
characteristic curves and precision recall curves along with
accumulation curves, per user request. Finally, more modeling methods
will be added as we identify those with appeal to our user base. Support
for a particular model may be requested here:
\url{https://github.com/jrash/chemmodlab/issues}.

\section{Aknowledgments}\label{aknowledgments}

The authors would like to than Denis Fourches for his helpful comments.

\bibliography{ash-hughes-oliver}

\begin{thebibliography}{17}
\providecommand{\natexlab}[1]{#1}
\providecommand{\url}[1]{\texttt{#1}}
\expandafter\ifx\csname urlstyle\endcsname\relax
  \providecommand{\doi}[1]{doi: #1}\else
  \providecommand{\doi}{doi: \begingroup \urlstyle{rm}\Url}\fi

\bibitem[Beygelzimer et~al.(2013)Beygelzimer, Kakadet, Langford, Arya, Mount,
  and Li]{Beygelzimer2013}
A.~Beygelzimer, S.~Kakadet, J.~Langford, S.~Arya, D.~Mount, and S.~Li.
\newblock \emph{FNN: Fast Nearest Neighbor Search Algorithms and Applications},
  2013.
\newblock URL \url{https://CRAN.R-project.org/package=FNN}.
\newblock R package version 1.1.

\bibitem[Brin and Page(1998)]{Brin1998}
S.~Brin and L.~Page.
\newblock {The Anatomy of a Large-Scale Hypertextual Web Search Engine}.
\newblock \emph{Computer Networks and ISDN Systems}, 30\penalty0
  (1-7):\penalty0 107--117, 1998.
\newblock ISSN 01697552.
\newblock URL \url{https://doi.org/10.1016/s0169-7552(98)00110-x}.

\bibitem[Burden(1989)]{Burden1989}
F.~R. Burden.
\newblock {Molecular Identification Number for Substructure Searches}.
\newblock \emph{Journal of Chemical Information and Modeling}, 29\penalty0
  (3):\penalty0 225--227, 1989.
\newblock ISSN 1549-9596.
\newblock URL \url{https://doi.org/10.1021/ci00063a011}.

\bibitem[Cherkasov et~al.(2014)Cherkasov, Muratov, Fourches, Varnek, Baskin,
  Cronin, Dearden, Gramatica, Martin, Todeschini, Consonni, Kuz'min, Cramer,
  Benigni, Yang, Rathman, Terfloth, Gasteiger, Richard, and
  Tropsha]{Cherkasov2014}
A.~Cherkasov, E.~N. Muratov, D.~Fourches, A.~Varnek, I.~I. Baskin, M.~Cronin,
  J.~Dearden, P.~Gramatica, Y.~C. Martin, R.~Todeschini, V.~Consonni, V.~E.
  Kuz'min, R.~Cramer, R.~Benigni, C.~Yang, J.~Rathman, L.~Terfloth,
  J.~Gasteiger, A.~Richard, and A.~Tropsha.
\newblock {QSAR Modeling: Where Have You Been? Where Are You Going To?}
\newblock \emph{Journal of Medicinal Chemistry}, 57\penalty0 (12):\penalty0
  4977--5010, 2014.
\newblock ISSN 0022-2623.
\newblock URL \url{https://doi.org/10.1021/jm4004285}.

\bibitem[from Jed~Wing et~al.(2016)from Jed~Wing, Weston, Williams, Keefer,
  Engelhardt, Cooper, Mayer, Kenkel, the R~Core~Team, Benesty, Lescarbeau,
  Ziem, Scrucca, Tang, Candan, and Hunt.]{Kuhn2016}
M.~K.~C. from Jed~Wing, S.~Weston, A.~Williams, C.~Keefer, A.~Engelhardt,
  T.~Cooper, Z.~Mayer, B.~Kenkel, the R~Core~Team, M.~Benesty, R.~Lescarbeau,
  A.~Ziem, L.~Scrucca, Y.~Tang, C.~Candan, and T.~Hunt.
\newblock \emph{Caret: Classification and Regression Training}, 2016.
\newblock URL \url{https://CRAN.R-project.org/package=caret}.
\newblock R package version 6.0-73.

\bibitem[Hastie et~al.(2009)Hastie, Tibshirani, and Friedman]{Hastie2009}
T.~Hastie, R.~Tibshirani, and J.~Friedman.
\newblock \emph{{The Elements of Statistical Learning}}, volume~1 of
  \emph{Springer Series in Statistics}.
\newblock Springer-Verlag, 2009.
\newblock ISBN 978-0-387-84857-0.
\newblock URL \url{https://doi.org/10.1007/b94608}.

\bibitem[Hughes-Oliver et~al.(2011)Hughes-Oliver, Brooks, Welch, Khaledi,
  Hawkins, Young, Patil, Howell, Ng, and Chu]{Hughes-Oliver2011}
J.~M. Hughes-Oliver, A.~D. Brooks, W.~J. Welch, M.~G. Khaledi, D.~Hawkins,
  S.~S. Young, K.~Patil, G.~W. Howell, R.~T. Ng, and M.~T. Chu.
\newblock {ChemModLab: a Web-Based Cheminformatics Modeling Laboratory.}
\newblock \emph{In silico biology}, 11\penalty0 (1-2):\penalty0 61--81, 2011.
\newblock ISSN 1434-3207.
\newblock URL \url{https://doi.org/10.3233/ci-2008-0016}.

\bibitem[James et~al.(2013)James, Witten, Hastie, and Tibshirani]{James2013}
G.~James, D.~Witten, T.~Hastie, and R.~Tibshirani.
\newblock \emph{{An Introduction to Statistical Learning}}, volume 103 of
  \emph{Springer Texts in Statistics}.
\newblock Springer-Verlag, 2013.
\newblock ISBN 978-1-4614-7137-0.
\newblock URL \url{https://doi.org/10.1007/978-1-4614-7138-7}.

\bibitem[Kearsley et~al.(1996)Kearsley, Sallamack, Fluder, Andose, Mosley, and
  Sheridan]{Kearsley}
S.~K. Kearsley, S.~Sallamack, E.~M. Fluder, J.~D. Andose, R.~T. Mosley, and
  R.~P. Sheridan.
\newblock {Chemical Similarity Using Physiochemical Property Descriptors}.
\newblock \emph{Journal of Chemical Information and Modeling}, 36:\penalty0
  118--127, 1996.
\newblock ISSN 1549-9596.
\newblock URL \url{https://doi.org/10.1021/ci950274j}.

\bibitem[Kim(2009)]{Kim2009}
J.-H. Kim.
\newblock {Estimating Classification Error Rate: Repeated Cross-Validation,
  Repeated Hold-out and Bootstrap}.
\newblock \emph{Computational Statistics {\&} Data Analysis}, 53\penalty0
  (11):\penalty0 3735--3745, 2009.
\newblock ISSN 01679473.
\newblock URL \url{https://doi.org/10.1016/j.csda.2009.04.009}.

\bibitem[Kramer(1956)]{Kramer1956}
C.~Y. Kramer.
\newblock {Extension of Multiple Range Tests to Group Means with Unequal
  Numbers of Replications}.
\newblock \emph{Biometrics}, 12\penalty0 (3):\penalty0 307, 1956.
\newblock ISSN 0006341X.
\newblock URL \url{https://doi.org/10.2307/3001469}.

\bibitem[Kuhn(2008)]{McCollum2009}
M.~Kuhn.
\newblock {Building Predictive Models in R Using the Caret Package}.
\newblock \emph{Journal of Statistical Software}, 28\penalty0 (5):\penalty0
  159--160, 2008.
\newblock ISSN 1548-7660.
\newblock URL \url{https://doi.org/10.18637/jss.v028.i05}.

\bibitem[Kuhn and Johnson(2013)]{Kuhn2013}
M.~Kuhn and K.~Johnson.
\newblock \emph{{Applied Predictive Modeling}}.
\newblock Springer-Verlag, New York, NY, 2013.
\newblock ISBN 978-1-4614-6848-6.
\newblock URL \url{https://doi.org/10.1007/978-1-4614-6849-3}.

\bibitem[Liu et~al.(2005)Liu, Feng, and Young]{Liu}
K.~Liu, J.~Feng, and S.~S. Young.
\newblock {PowerMV: A Software Environment for Molecular Viewing, Descriptor
  Generation, Data Analysis and Hit Evaluation}.
\newblock \emph{Journal of Chemical Information and Modeling}, 45\penalty0
  (2):\penalty0 515--522, 2005.
\newblock ISSN 1549-9596.
\newblock URL \url{https://doi.org/10.1021/ci049847v}.

\bibitem[Molinaro et~al.(2005)Molinaro, Simon, and Pfeiffer]{Molinaro2005}
A.~M. Molinaro, R.~Simon, and R.~M. Pfeiffer.
\newblock {Prediction Error Estimation: a Comparison of Resampling Methods}.
\newblock \emph{Bioinformatics}, 21\penalty0 (15):\penalty0 3301--3307, 2005.
\newblock ISSN 1367-4803.

\bibitem[Shen et~al.(2011)Shen, Welch, and Hughes-Oliver]{Shen2011}
H.~Shen, W.~J. Welch, and J.~M. Hughes-Oliver.
\newblock {Efficient, Adaptive Cross-Validation for Tuning and Comparing
  Models, with Application to Drug Discovery}.
\newblock \emph{The Annals of Applied Statistics}, 5\penalty0 (4):\penalty0
  2668--2687, 2011.
\newblock ISSN 1932-6157.
\newblock URL \url{https://doi.org/10.1214/11-aoas491}.

\bibitem[Tukey and Berringer(1994)]{Tukey1994}
J.~W. Tukey and D.~R. Berringer.
\newblock \emph{{The Collected Works of John W. Tukey: Multiple Comparions,
  Volume VIII}}.
\newblock Chapman {\&} Hall/CRC, Princeton, 8th edition, 1994.
\newblock ISBN 0412051214.

\end{thebibliography}

\address{%
Jeremy R. Ash\\
North Carolina State University\\
Bioinformatics Research Center\\ Department of Statistics\\ 335 Ricks Hall\\ Campus Box 7566\\ Raleigh, NC 27695-8203, USA\\ ORCiD: 0000-0002-8041-8524\\
}
\href{mailto:jrash@ncsu.edu}{\nolinkurl{jrash@ncsu.edu}}

\address{%
Jacqueline M. Hughes-Oliver\\
North Carolina State University\\
Department of Statistics\\ 2311 Stinson Drive\\ Campus Box 8203\\ Raleigh, NC 27695-8203, USA\\
}
\href{mailto:hughesol@ncsu.edu}{\nolinkurl{hughesol@ncsu.edu}}

\end{article}

\end{document}